# Computational Algorithms Based on the Paninian System to Process Euphonic Conjunctions for Word Searches

Kasmir Raja S. V.
Dean – Research
SRM University
Chennai, India

Rajitha V.
Department of Computer Science
Meenakshi College for Women
Chennai, India
&
Research Scholar
Mother Teresa Women's University
Kodaikanal, India

Meenakshi Lakshmanan
Department of Computer Science
Meenakshi College for Women
Chennai, India
meenakshi.lakshmanan@gmail.com

*Abstract* – **Searching for words in Sanskrit E-text is a problem that is accompanied by complexities introduced by features of Sanskrit such as euphonic conjunctions or 'sandhis'. A word could occur in an E-text in a transformed form owing to the operation of rules of** sandhi**. Simple word search would not yield these transformed forms of the word. Further, there is no search engine in the literature that can comprehensively search for words in Sanskrit E-texts taking euphonic conjunctions into account. This work presents an optimal binary representational schema for letters of the Sanskrit alphabet along with algorithms to efficiently process the** sandhi **rules of Sanskrit grammar. The work further presents an algorithm that uses the** sandhi **processing algorithm to perform a comprehensive word search on E-text.**

## I. INTRODUCTION

Word search in Sanskrit E-texts is a problem that is beset with complexities, unlike in the case of English E-texts. The problem assumes relevance in the context of the availability of rapidly increasing numbers of ancient Sanskrit texts [5-9] in the electronic format. The importance of n-gram analysis of Sanskrit texts for scholars and the tremendous utility of locating specific words in a variety of texts to aid the scholastic process can hardly be overemphasized.

Dating of a text, fixing its authorship with certainty, and analysis of the writing style of an author of a text, are some of the areas in which n-grams assume criticality especially in the context of ancient Sanskrit works. Quoting from authoritative texts is imperative in scholarly works, and word searches can provide crucial help in this regard. Locating the portion in a text or texts in which a particular usage or word is found is of great importance to scholars who write explanatory treatises of Sanskrit-based works in English and other languages. Semantic analysis and understanding of texts are facilitated by finding occurrences of words and studying them in different contexts. In fact, ancient Sanskrit works are universally acknowledged as being mines of information on a whole spectrum of disciplines, and hence finding actual occurrences of words is of great consequence to not only Sanskrit scholars but to also researchers from various other disciplines ranging from philosophy, theology, the arts and the physical and life sciences to sociology, medicine and astronomy.

## II. THE PROBLEM

As stated above, there are complexities involved in searching comprehensively for words in a Sanskrit E-text. One of the major contributors to this complexity is the operation of euphonic conjunctions or '*sandhis*'. A *sandhi* is a point in a word or between words, at which adjacent letters coalesce and transform [3]. This is a common feature in many Indian languages as against European languages, and has far-reaching consequences in Sanskrit. The transformation caused by the application of rules of *sandhi* in Sanskrit can be significant enough to alter the word itself to such a degree that the transformed word would not show up in a simple word search.

For example, the word '*asamardhiḥ*' (meaning of unmatched affluence), can be transformed into '*āsamardhiḥ*' because of the operation of a euphonic conjunction with a word ending in '*a*' preceding it, or '*āsamardhir*' or '*āsamardhis*' in combination with words occurring after it or '*asamarddhiḥ*' or '*asamardddhiḥ*' by internal transformation. Clearly, simply searching for the word *asamardhiḥ* would not yield the occurrences of the same word as '*asamarddhir*', '*āsamardddhir*', or other alternative forms. As such, a normal text-search using a Unicode text editor would not suffice. Other search engines currently used for Sanskrit [13] too do not provide for such comprehensive searching.

In order to achieve such an exhaustive search, all possible forms of the word resulting from the euphonic conjunctions that would become operative in its case must be generated and searched for in the given text.

The authors have already presented a new schema for fast *sandhi* processing in earlier work [4]. The present work extends the application of that schema to other *sandhi* rules including consonant-based and *visarga*-based *sandhis* as well as important rules with respect to exceptional cases. It further presents a complete computational algorithm to process all *sandhis*, and an algorithm to apply this *sandhi*-processing procedure to generate all word forms to enable comprehensive searching.

*A. Language Representation*

The Unicode hexadecimal range 0900 - 097F is used to represent Sanskrit characters in *Devanāgarī* script. The characters used to represent Sanskrit letters in English script are found in the Basic Latin (0000-007F), Latin-1 Supplement (0080-00FF), Latin Extended-A (0100-017F) and Latin Extended Additional (1E00 – 1EFF) Unicode ranges.

The Latin character set has been employed in this work to represent Sanskrit letters as E-text. As such, the schema and algorithms presented do not use *Devanāgarī* script. To use the algorithms for text that is in *Devanāgarī* script, the text needs to first be converted to Latin text.

*B. Terminology*

The terminology employed in this work for certain groups of letters of the Sanskrit alphabet is given in Table 1.

**Table 1: Terminology**

| Term | Description / Notation |
|---|---|
| Vowel | *a,ā,i,ī,u,ū,ṛ,ṝ,ḷ,e,ai,o,au* |
| Semi-vowel | *y, v, r, l* |
| Consonant | *k,kh,g,gh,ṅ,c,ch,j,jh,ñ,ṭ,ṭh,ḍ,ḍh,ṇ,t,th,d,dh, n,p,ph,b,bh,m,ś,ṣ,s,h* |
| Guttural | *k, kh, g, gh, ṅ* |
| Palatal | *c, ch, j, jh, ñ* |
| Cerebral | *ṭ, ṭh, ḍ, ḍh, ṇ* |
| Dental | *t, th, d, dh, n* |
| Labial | *p, ph, b, bh, m* |
| Nasal | *ṅ, ñ, ṇ, n, m* |
| Aspirate | *h* |
| Sibilant | *ś, ṣ, s* |
| Column1 | *k, c, ṭ, t, p* |
| Column2 | *kh, ch, ṭh, th, ph* |
| Column3 | *g, j, ḍ, d, b* |
| Column4 | *gh, jh, ḍh, d, bh* |
| *Visarga* | *ḥ* |
| *Anusvāra* | *ṁ* |
| Hard consonant | *Column1, Column2, Sibilants* |
| Soft consonant | *Column3, Column4, Nasals, Aspirate* |
| Hard guttural | *k, kh* |
| Hard labial | *p, ph* |
| Mutes | *Column1, Column2, Column3, Column4, Nasals* |
| *Jihvāmūlīya* | ʌ (pronounced as the end of '*kah*') |
| *Upadhmānīya* | ɣ (pronounced as the end of '*paf*') |

### III. THE BASIS OF THE WORK

The renowned ancient Sanskrit linguist, Pāṇini, codified the extant grammar of Sanskrit into terse aphorisms ('*sūtras*') and organized these aphorisms into eight chapters. This work is the authoritative *Aṣṭādhyāyī* (literally meaning 'work in eight chapters') and is universally acknowledged as the most comprehensive codification of the grammar of any language. The grammatical rules that make up Pāṇini's *Aṣṭādhyāyī* are derivational and known for their mathematical precision in spite of dealing with the nuances of the language at various levels including morphology, syntax, semantics, phonology and pragmatics. Owing to the cryptic nature of the *Aṣṭādhyāyī*, one or more of the commentaries on it are required to get a clear understanding of its contents.

The current work deals with Pāṇini's *sandhi*-related aphorisms with the help of the recognized commentaries, *Siddhānta-kaumudī* [1] and *Kāśikā* [2]. Both these commentaries are accepted by Sanskrit scholars as authoritative works on Pāṇinian grammar.

Pāṇini's statements of grammatical rules are expressed on the basis of the *Māheśvara-sūtras*, or the 'aphorisms of Maheśvara'. These aphorisms provide a list of all the letters of the Sanskrit alphabet ordered in a specific sequence. The *Māheśvara* aphorisms are given below:

1. *a-i-u-ṇ*
2. *ṛ-ḷ-k*
3. *e-o-ṅ*
4. *ai-au-c*
5. *ha-ya-va-ra-ṭ*
6. *la-ṇ*
7. *ña-ma-ṅa-ṇa-na-m*
8. *jha-bha-ñ*
9. *gha-ḍha-dha-ṣ*
10. *ja-ba-ga-ḍa-da-ś*
11. *kha-pha-cha-ṭha-tha-ca-ṭa-ta-v*
12. *ka-pa-y*
13. *śa-ṣa-sa-r*
14. *ha-l*

The last letter in each of these aphorisms is only a place-holder. The first four aphorisms list only the short forms of all the vowels, while the rest list the semi-vowels and consonants; the latter list has the vowel '*a*' appended to each letter only to enable pronunciation of the aphorism.

*A. The Approach*

The present work is based on earlier work by the authors, which directly codifies Pāṇini's rules in a novel way using binary representations [4]. The unique data representation devised by the authors has been further refined in this work and consonant-based, *visarga*-based *sandhi* rules, as well as some special *sandhi* rules have been included in this work.

Rule representation has been simplified to minimal binary set-unset operations. Further, the *sūtra* ordering has been done after acquiring a thorough understanding of the operation of Pāṇini's *sandhi*-related aphorisms. As such, this work presents a significant extension, refinement and closure of the earlier work of the authors. Moreover, it provides a clear understanding of the rules governing *sandhi* as laid down by Pāṇini, in a comprehensive and simplified way, hitherto not encountered in the literature.

*B. The Binary Schema*

The following is an extract from already published work by the authors [4] and is included here for completeness of the presentation.

A point of *sandhi* is denoted by

$$x + y$$

where $x$ and $y$ denote the *sandhi* letters and the symbol '+' denotes adjacency. The variable $X$ denotes the sequence of letters culminating in $x$; the variable $Y$ denotes the sequence of letters starting with $y$. The notations $X$ and $Y$ are used to depict special conditions that pertain to an entire word or sequence of letters involved in the *sandhi* rule. The letter immediately preceding $x$ and the letter immediately succeeding $y$ are denoted respectively as $u$ and $w$ respectively.

The refined schematic developed in this work to represent letters of the Sanskrit alphabet is given in Table 2.

**Table 2: Binary representation scheme**

| # | Letters |
|---|---|
| 0 | a,ā,i,ī,u,ū,ṛ,ṝ,ḷ,e,ai,o,au |
| 1 | y,r,l,v,yaṁ,vaṁ,laṁ |
| 2 | k,kh,g,gh,ṅ,c,ch,j,jh,ñ,ṭ,ṭh,ḍ,ḍh,ṇ,t,th,d,dh,n,p,ph,b,bh,m,ś,ṣ,s,h |
| 3 | ṁ,ḥ,',#,x,f |
| 4 | a |
| 5 | ā |
| 6 | u,i |
| 7 | ū,ī |
| 8 | u,i,ṛ,ḷ,a |
| 9 | ū,ī,ṝ,ṝ,ā |
| 10 | u,i,ṛ,ḷ |
| 11 | ū,ī,ṝ |
| 12 | ṛ,ṝ,ḷ |
| 13 | o,e |
| 14 | au,ai |
| 15 | o,au,e,ai |
| 16 | o,e,ar,al |
| 17 | ār,ār,āl |
| 18 | av,āv,ay,āy |
| 19 | ava |
| 20 | v,y,r,l |
| 21 | r |
| 22 | ṁv,ṁy,r,ṁl |
| 23 | ṣ |
| 24 | s |
| 25 | ś |
| 26 | h |
| 27 | ṅ,ṇ,n,ñ,m |
| 28 | ṅ,ṇ |
| 29 | n |
| 30 | m |
| 31 | k,gh,kh,g,ṅ |
| 32 | c,jh,ch,j,ñ,ś |
| 33 | ṭ,ḍh,ṭh,ḍ,ṇ,ṣ |
| 34 | t,dh,th,d,n,s |
| 35 | p,bh,ph,b,m |
| 36 | k,ṭ,t,c,p |
| 37 | kh,ṭh,th,ch,ph |
| 38 | g,ḍ,d,j,b |
| 39 | gh,ḍh,dh,jh,bh |
| 40 | ch,ṭh,th,c,ṭ,t |
| 41 | k,kh,p,ph |
| 42 | x,x,f,f |
| 43 | ṁ |
| 44 | ḥ |
| 45 | ' |
| 46 | # |

In Table 2, $x$ is the *jihvāmūlīya* and $f$ is the *upadhmānīya* mentioned in Table 1; the # symbol stands for '*ru*' a special intermediary form of the semi-vowel *r*.

Any letter of the alphabet is represented in two parts: Part 1 denotes the category to which a letter belongs (zero-based serial number in Table 2), and Part 2 denotes the zero-based term number within the series that the letter is or fits into. In any letter representation, Part 1 is a binary string of fixed length 46, in which the set bit denotes the category number, while Part 2 is a binary string of maximum length 6 in which the set bit indicates which particular letter is being represented. It is clear that one letter has many representations under this scheme.

The first four shaded rows of Table 2 stand for overall categories, viz. vowels, semi-vowels, consonants and special characters respectively. One of these four bits have to be set in any letter representation. There is no corresponding Part 2 value for the bits 0, 1, 2 and 3 of Part 1.

For simplicity of presentation, *sandhi* rules use the following notation: $x_i(n) = 1$ indicates that the $n$th bit of Part $i$ of the variable $x$ is set, where $i = 1, 2$. In the implementation, the checks for bit set can be done by simply using the XOR operation.

IV. *SANDHI* PROCESSING UNDER THE PĀṆINIAN SYSTEM

Each of the eight chapters of Pāṇini's *Aṣṭādhyāyī* is divided into four parts or *pādas*. Overall, the work is defined by Pāṇini as consisting of two parts, the *sapādasaptādhyāyī* (the aphorisms of Chapters 1.1 to 8.1), and the *tripādī* (the aphorisms of Chapters 8.2 to 8.4).

The order in which the rules should be visited was arrived at in this work after a thorough study of Pāṇini's aphorisms with respect to euphonic conjunctions. As a result of the study, the set of *sandhi* rules has been split into two in this work: Set 1, having all the relevant aphorisms of the *sapādasaptādhyāyī* as well as a few specific rules from the *tripādī*, and Set 2, having all the remaining relevant aphorisms of the *tripādī*.

The order of parsing is as follows: rules in Set 1 are parsed in reverse order of their *Aṣṭādhyāyī* order; rules in Set 2 are parsed in the *Aṣṭādhyāyī* order itself. (The *sūtra* number as it appears in the *Aṣṭādhyāyī* is indicated in the algorithm between double pipe symbols given after the *sūtra*.)This parsing order is adopted so that no rule already parsed has to be parsed again. As such, the flow of the program is just from top to bottom. There are exceptions to the above parsing orders in both sets that arise because of certain overruling *sūtras* that appear earlier / later respectively in the two sets. The ordering is changed to accommodate such rules in such a way as to parse them before the main rule.

Assuming that the rules are ordered in the above manner in the two sets, the following general algorithm for parsing rules is presented. The *word_list* is a list of the alternative word-pair outputs generated, and represents the output at the end of the algorithm.

Algorithm SandhiRulesParser
{
    while Set 1 rules have not been fully visited
    {
        Try the next rule in Set 1;
        if the rule applies
        {
            Apply the rule and store the output;
            if the rule is optional
            {
                Add the current word pair to the *word_list*;
                Continue checking from the next rule for all word pairs stored in the *word_list*;
            }
            else
            {
                Process internal *sandhi* rules of Set 2;
            }
        }
    }
    while Set 2 rules have not been fully visited
    {
        Try the next rule in Set 2;
        if the rule applies
        {
            Apply the rule and transform the given words;
            if the rule is optional
            {
                Add the current word pair to the *word_list*;
                Continue checking from the next rule for all word pairs stored in the *word_list*;
            }
        }
    }
}

There are a few exceptions that would apply to the general processing order prescribed by the above algorithm. For example, in Set 1 the output produced by an optionally applying rule does, in certain cases, have to pass through a rule appearing below and undergo further transformation as a result, as it happens in Set 2. Also, it is found that a few rare rules of Set 2 have to be processed before Set 1. Further, in Set 2, all rules that form exceptions to a particular rule are stated after it by Pāṇini, but clearly, have to be processed before the rule by the algorithm.

## V. THE SANDHI PROCESSOR

The key to symbols used in rule coding and algorithm specification is as follows:

- // means single-line explanatory comment
- { } are block or set indicators
- ∧ denotes *and*
- ∨ denotes *or*
- ¬ denotes *not*
- ⊕ denotes *xor*
- | denotes word concatenation

The algorithm SandhiProcessor processes the rules pertaining to all the major *sandhis* in Sanskrit grammar, in accordance with the processing scheme presented in Algorithm SandhiRulesParser. Set 1 and Set2 *sandhis* have been incorporated here one below the other and the rules have been codified as per the schema presented above. The vowel *sandhis* presented in [4] have been modified in accordance to the reduced schema and included here for completeness.

When $x_i$, $y_i$, etc., for $i = 1,2$, are assigned new values by setting bits, it is assumed that their initial values are first unset. Also, if either part of a variable is not set, it is assumed to remain unchanged. Further, it is also assumed that a category change caused by a *sandhi* will cause an automatic change in the first four bits of the letter representation and that all bit representations for the changed letter will be generated thereafter. Hence, these aspects are not explicitly stated for each rule in this algorithm.

The speed of processing is increased by going into a rule only if overall conditions are satisfied. For instance, if the rule is a vowel *sandhi* rule, where both $x$ and $y$ are required to be vowels, then the check if $x_1(0)$ and $y_1(0)$ are 1 is first made. If this bit-check is not true for the input words, then a whole set of vowel *sandhis* is omitted from the parse, thus increasing the efficiency of the algorithm. These overall checks have not been shown in the algorithm presented below, in order to make the presentation more simple.

Algorithm SandhiProcessor (*X*,*Y*)
{
    //1.*svaujasamauṭchaṣṭābhyāṁbhisṅebhyāṁbhyasṅasi
    //bhyāṁbhyasṅasosāmṅyossup* || 4.1.2 ||
    //If there is a *visarga* (*ḥ*) at the end of *X*, then the *visarga* is
    //changed to '*s*'.
    if $x_1(44)$
    {
        $x_1(24) = 1$;
    }

    //2. *sasajuṣo ruḥ* || 8.2.66 ||
    // Common name: *visarga-rutva sandhi*
    //If last letter of *X* is *s*, then s is replaced by '#' which
    //stands for the particle '*ru*', interpreted as '*r*'.
    //This rule is incorporated here though it belongs to the
    //*sapādasaptādhyāyī* .
    if $x_1(24)$
    {
        $x_1(46) = 1$;
    }

    //3. *avaṅ sphoṭāyanasya* || 6.1.123 || (vowel *sandhi*)
    // Common name: *avaṅādeśa sandhi*
    //If the word *go* is followed by a vowel, then the *o* of *go*
    //is optionally replaced by *ava*.
    if $(x_1(15) \land x_2(0)) \land (u_1(31) \land u_2(3))$
    {
        Add *X*|*Y* to *word_list*;
        $x_1(19) = 1$;
    }

//4. *haśi ca* || 6.1.114 ||
//If # (*ru*) or *r* at the end of X is preceded by the vowel *a*
//and followed by aspirate, semi-vowel, nasal, Column3 or
//Column4, then last letter of X is replaced with the vowel
//'*u*' and shifted to Y to become the first letter of Y
$if\ (x_1(46) \vee x_1(21)) \wedge u_1(4) \wedge (y_1(1) \vee y_1(26) \vee y_1(27) \vee y_1(38) \vee y_1(39))$
{
    $x_1(6) = 1;$
    $x_2 = u_2;$
    Shift $x$ from the end of $X$ to the beginning of $Y$;
}

//5. *ato roraplutādaplute* || 6.1.113 ||
//Common name: *visarga-rutva sandhi*
//If # (*ru*) or *r* at the end of X is followed and preceded by
//*a*, then the # or *r* is replaced with the vowel '*u*' and shifted
//to Y to become the first letter of Y
$if\ (x_1(46) \vee x_1(21)) \wedge u_1(4) \wedge y_1(4)$
{
    $x_1(6) = 1;$
    $x_2 = u_2;$
    Shift $x$ from the end of $X$ to the beginning of $Y$;
}

//6. *eṅaḥ padāntādati* || 6.1.109 || (vowel *sandhi*)
// Common name: *pūrvarūpa sandhi*
//If *e* or *o* at the end of a word is followed by *a*, then *e* or *o*
//remains, and the *avagraha* (') replaces *a*.
$if\ x_1(13) \wedge y_1(4)$
{
    $y_1(45) = 1;$
}

//7. *akaḥ savarṇe dīrghaḥ* || 6.1.101 || (vowel *sandhi*)
//Common name: *savarṇadīrgha sandhi*
//If one of *a, i, u, ṛ* or *ḷ* or their long equivalents *ā, ī, ū* and *ṝ*
//is followed by the short or long form of the same letter,
//then the corresponding long letter replaces both.
$if\ (x_1(8) \vee x_1(9)) \wedge (y_1(8) \vee y_1(9)) \wedge \neg(x_2 \oplus y_2)$
{
    delete $y$;
    $x_1(9) = 1;$
    return $X|Y$;
}

//8. *omāṅośca* || 6.1.95 || (vowel *sandhi*)
// Common name: *pararūpa sandhi*
//If *a* or *ā* is followed by *o* of the word *om* or *oṁ*, then *o*
//replaces both.
$if\ (x_1(4) \vee x_1(5)) \wedge Y \in \{om, oṁ\}$
{
    delete $x$;
}

//9. *etyedhatyūṭhsu* || 6.1.89 || (vowel *sandhi*)
//Common name: *vṛddhi sandhi*
//For this rule, in all cases the resultant letter replaces $x$
//and $y$.
//i) If *a* or *ā* is followed by *eti* or *edhati*, then *vṛddhi* letter
//*ai* replaces both
//ii) If the preposition *pra* is followed by *eṣa* or *eṣy*, then
//*vṛddhi* letter *ai* replaces both
//iii) If *a* or *ā* is followed by *ūh*, then *vṛddhi* letter *au*
//replaces both
//iv) If preposition *pra* is followed by *ūḍh*, then *vṛddhi*
//letter *au* replaces both
//v) If word *sva* is followed by *īr*, then *vṛddhi* letter *ai*
//replaces both
$if\ (x_1(4) \vee x_1(5))$ //$x$ is '*a*' or '*ā*'
{
    $if\ (y_1(13) \wedge y_2(1))$ //$y$ is '*e*'
    {
        $if$ Y starts with {*et, edhat*} //(i)
        {
            delete $x$;
            $y_1(14) = 1;$
        }
        $elseif\ X$ = '*pra*' $\wedge\ Y$ starts with {*eṣ, eṣy*} //(ii)
        {
            delete $x$;
            $y_1(14) = 1;$
        }
    }
    $elseif\ (y_1(7) \wedge y_2(0))$ //$y$ is '*ū*'
    {
        $if\ w_1(26)$ // (iii)
        {
            delete $x$;
            $y_1(14) = 1;$
        }
        $elseif\ X$ = '*pra*' $\wedge\ Y$ starts with {*ūḍh*} //(iv)
        {
            delete $x$;
            $y_1(14) = 1;$
        }
    }
    $elseif\ X$ = '*sva*' $\wedge\ (y_1(7) \wedge y_2(1)) \wedge w_1(21)$//(v)
    {
        delete $x$;
        $y_1(14) = 1;$
    }
}

//10. *eṅi pararūpaṁ* || 6.1.94 ||   (vowel *sandhi*)
// Common name: *pararūpa sandhi*
//If *a* or *ā* at the end of a preposition is followed by *e* or
//*o*, then the *e* or *o* replaces both.
//Note: The prepositions that qualify are: *pra, ava, apa,*
//*upa, parā*.
$if\ X \in \{pra, ava, apa, upa, parā\} \wedge y_1(13)$
{
    delete $x$;
}

//11. *upasargādṛti dhātau* || 6.1.91 || (vowel *sandhi*)
//Common name: *vṛddhi sandhi*
//i) If *a* or *ā* at the end of a preposition is followed by *ṛ*,

//ṝ or ḷ, then *vṛddhi* letter *ār, ār* or *āl* respectively
//replaces both. *Note:* The prepositions that qualify are:
//*pra, parā, apa, ava, upa*
//ii) If the word *vatsara, kambala, vasana, daśa, ṛṇa* is
//followed by the word *ṛṇa*, then *vṛddhi* letter *ār*
//replaces both.
//Note: This rule clashes with 6.1.87 (*guṇa sandhi*), and
//takes precedence.
$if\ X\ \in \{pra, ava, apa, upa, parā\} \wedge y_1(12)$
{
    delete $x$;
    $y_1(17) = 1$;
}
$if\ X\ \in \{vatsara, kambala, vasana, daśa, ṛṇa\} \wedge$
$(y_1(12) \wedge y_2(0)) \wedge Y = \text{'}ṛṇa\text{'}$
{
    delete $x$;
    $y_1(17) = 1$;
}

//12. *vṛddhireci* || 6.1.88 || (vowel *sandhi*)
// Common name: *vṛddhi sandhi*
//If *a* or *ā* is followed by *e, o, ai* or *au*, then the
//corresponding *vṛddhi* letter *ai* or *au* replaces both.
$if\ (x_1(4) \vee x_1(5)) \wedge ((y_1(13) \vee y_1(14))$
{
    delete $x$;
    $y_1(14) = 1$;
}

//13. *ādguṇaḥ* || 6.1.87 || (vowel *sandhi*)
//    *uraṇ raparaḥ* || 1.1.51 ||
// Common name: *guṇa sandhi*
//If *a* or *ā* is followed by *i, ī, u, ū, ṛ, ṝ* or *ḷ*, then the
//corresponding *guṇa* letter *e, o, ar* or *al* replaces both.
$if\ (x_1(4) \vee x_1(5)) \wedge ((y_1(10) \vee y_1(11))$
{
    delete $x$;
    $y_1(16) = 1$;
}

//14. *ecoyavāyāvaḥ* || 6.1.78 || (vowel *sandhi*)
// Common name: *ayāyāvādeśa sandhi*
//If *e, o, ai* or *au* is followed by a vowel, then *ay, av, āy,*
//*āv* replace the first respectively.
$if\ x_1(15) \wedge y_1(0)$
{
    $x_1(18) = 1$;
}

//15. *iko yaṇaci* || 6.1.77 || (vowel *sandhi*)
// Common name: *yaṇādeśa sandhi*
//If *i, ī, u, ū, ṛ, ṝ* or *ḷ* is followed by a vowel, then the
//corresponding semi-vowel (*y, v, r, l*) replaces the first.
$if\ (x_1(10) \vee x_1(11)) \wedge y_1(0)$
{
    $x_1(20) = 1$;
}

//16. *che ca* || 6.1.73 ||
//Common name: *tugāgama sandhi*
//If a short vowel is followed by the consonant *ch*, then
//*t* is added.
$if\ x_1(8) \wedge (y_1(40) \wedge y_2(0))$
{
    $z_1(34) = 1$;
    $z_2 = y_2$;
    Add z to the end of $X$;
}

//17. *āṅmāṅośca* || 6.1.74 ||
//Common name: *tugāgama sandhi*
//If the particle *ā* or word *mā* is followed by *ch*,
//then *t* is added.
$if\ X \in \{ā, mā\} \wedge (y_1(40) \wedge y_2(0))$
{
    $z_1(34) = 1$;
    $z_2 = y_2$;
    Add z to the end of $X$;
}

//18. *dīrghāt* || 6.1.75 ||
//    *padāntādvā* || 6.1.76 ||
//Common name: *tugāgama sandhi*
//If a long vowel is followed by *ch*, then *t* is added.
$if\ x_1(9) \wedge (y_1(40) \wedge y_2(0))$
{
    $z_1(34) = 1$;
    $z_2 = y_2$;
    Add z to the end of $X$;
}

//19. *saṁyogāntasya lopaḥ* || 8.2.23 ||
//If the final consonant of $X$ is preceded by a
//consonant, then the last consonant is dropped.
$if\ x_1(2) \wedge u_1(2)$
{
    delete $x$;
}

//20. *jhalāṁ jaśo'nte* || 8.2.39 ||
//Common name: *jaśtva sandhi*
//If $x$ is Column1, Column2, Column3, Column4, sibilant
//or aspirate, then $x$ is replaced by the corresponding
//Column3.
//Note: The rule *sasajuṣo ruḥ* || 8.2.66 || debars the
//application of this rule for words ending in sibilants, and
//has been incorporated earlier in Set 1 rules itself.
$if\ x_1(36) \vee x_1(37) \vee x_1(38) \vee x_1(39)$
{
    $x_1(38) = 1;$
}

//21. *pumaḥ khayyampare* ||8.3.6||
//    *atrānunāsikaḥ pūrvasya tu vā* || 8.3.2 ||
//If $X$ is the word '*pum*' or '*puṁ*' and is followed by

```
//Column1 or Column2, which is in turn followed by a
//vowel, semi-vowel or a nasal, then x is replaced by #
//(ru) and the preceding vowel is made nasal using the
//anusvāra.
if X ∈ {pum, pum̐} ∧ (y_1(36) ∨ y_1(37)) ∧ (w_1(0) ∨
w_1(1) ∨ w_1(27))
{
    x_1(43) = 1;
    z_1(46) = 1;
    x_2(0) = z_2(0) = 1;
}

//22. naśchavyapraśān || 8.3.7 ||
//     atrānunāsikaḥ pūrvasya tu vā || 8.3.2 ||
//Common name: satva sandhi
//If the final n of a word, except for the word praśān,
//followed by ch, ṭh, th, c, ṭ or t which is in turn
//followed by a vowel, semi-vowel or nasal, then
//n is replaced with # (ru) and the preceding vowel is
//made nasal using the anusvāra.
if ¬(X = 'praśān') ∧ (x_1(27) ∧ x_2(2)) ∧ y_1(40)) ∧
(w_1(0) ∨ w_1(1) ∨ w_1(27))
{
    u = x;
    x_1(43) = 1;
    z_1(46) = 1;
    x_2(0) = z_2(0) = 1;
}

//23. nr̥npe || 8.3.10 ||
//     atrānunāsikaḥ pūrvasya tu vā || 8.3.2 ||
//If X = 'nr̥n' and y is the letter 'p' then x is
//optionally replaced with # (ru) and the preceding
//vowel is made nasal using the anusvāra.
if X = 'nr̥n' ∧ (y_1(35) ∧ y_2(0))
{
    Add X|Y to word_list;
    u = x;
    x_1(43) = 1;
    z_1(46) = 1;
    x_2(0) = z_2(0) = 1;
}

//24. ro ri || 8.3.14 ||
//     ḍhralope pūrvasya dīrgho'ṇaḥ ||6.3.111||
//If r or # (ru) is followed by r and is preceded by a, i or u,
//then one r is dropped and the short vowel is made long.
if (x_1(46) ∨ x_1(21)) ∧ y_1(21)
{
    delete x;
    if u_1(6)
    {
        u_1(7) = 1;
    }
    if u_1(4)
    {
        u_1(5) = 1;
```

```
    }
}

//25. kharavasānayorvisarjanīyaḥ || 8.3.15 ||
//If x is # (ru) or r and is followed by a hard consonant,
//then x is replaced with visarga.
if (x_1(46) ∨ x_1(21)) ∧ (y_1(36) ∨ y_1(37) ∨ y_1(23) ∨
y_1(24) ∨ y_1(25))
{
    x_1(44) = 1;
}

//26. bhobhagoaghoapūrvasya yo'śi || 8.3.17 ||
//If x is # (ru) or r and is preceded by bho, bhago, agho,
//a or ā, and is followed by a vowel, semi-vowel or soft
//consonant, then x is replaced by the consonant 'y'.
if (x_1(46) ∨ x_1(21)) ∧ (X – {x}) in {bho, bhago, agho, a,
ā} ∧ (y_1(0) ∨ y_1(1) ∨ y_1(38) ∨ y_1(39) ∨ y_1(26) ∨
y_1(27))
{
    x_1(20) = 1;
    x_2(1) = 1;
}

//27. lopaḥ śākalyasya || 8.3.19 ||
//If the consonant 'y' or 'v' is preceded by a or ā and is
//followed by a vowel, semi-vowel or soft consonant,
//then the 'y' or 'v' is dropped.
if (x_1(20) ∧ (x_2(0) ∨ x_2(1))) ∧ (u_1(4) ∨ u_1(5)) ∧
(y_1(0) ∨ y_1(1) ∨ y_1(38) ∨ y_1(39) ∨ y_1(26) ∨ y_1(27))
{
    delete x;
}

//28. oto gārgyasya || 8.3.20 ||
//If the consonant 'y' is preceded by the vowel 'o' and
//followed by a vowel, semi-vowel or soft consonant,
//then the 'y' is dropped.
if (x_1(20) ∧ x_2(1)) ∧ (u_1(13) ∧ u_2(0)) ∧ (y_1(0) ∨
y_1(1) ∨ y_1(38) ∨ y_1(39) ∨ y_1(26) ∨ y_1(27))
{
    delete x;
}

//29. uñi ca pade || 8.3.21 ||
//If the consonant 'y' or 'v' is preceded by a or ā and is
//followed by the word 'u', then the 'y' or 'v' is dropped.
if (x_1(20) ∧ (x_2(0) ∨ x_2(1))) ∧ (u_1(4) ∨ u_1(5)) ∧ Y =
'u'
{
    delete x;
}

//30. hali sarveṣām̐ || 8.3.22 ||
//If the consonant 'y' is followed by a semi-vowel or
//consonant, then the 'y' is dropped.
```

```
if (x₁(20) ∧ x₂(1)) ∧ (y₁(1) ∨ y₁(2))
{
    delete x;
}
```

//31. *he mapare vā* || 8.3.26 ||
//i) If *m* is followed by *h* at the end of a word which is in
//turn followed by *m*, then the first *m* is optionally changed
//to *anusvāra*.
//ii) If *m* is followed by *h* which is in turn followed by
//consonants '*y*', '*l*', or '*v*', then the *m* is optionally replaced
//by the nasal forms of '*y*', '*l*', or '*v*' respectively.

```
if x₁(30) ∧ y₁(26)
{
    if w₁(30)
    {
        Add X|Y to word_list;
        x₁(43) = 1;
    }
    elseif w₁(20)
    {
        x₁(22) = 1;
        x₂ = w₂;
    }
}
```

//32. *napare naḥ* || 8.3.27 ||
//If *m* is followed by *h* at the end of a word which is in turn
//followed by *n*, then the *m* is optionally replaced by *n*.

```
if x₁(30) ∧ y₁(26) ∧ (w₁(27) ∧ w₂(2))
{
    Add X|Y to word_list;
    x = w;
}
```

//33. *naścāpadāntasya jhali* || 8.3.24 ||
//If *n* is followed by Column1, Column2, Column3,
//Column4, sibilant or aspirate not at the end of a word,
//then the *n* is replaced by *anusvāra*.

```
for (each letter x in a word and its succeeding letter y)
{
    if (x₁(29)) ∧ (y₁(36) ∨ y₁(37) ∨ y₁(38) ∨ y₁(39) ∨
                   y₁(23) ∨ y₁(24) ∨ y₁(25) ∨ y₁(26))
    {
        x₁(43) = 1;
    }
}
```

//34. *mo rāji samaḥ kvau* || 8.3.25 ||
//If *m* of the word '*sam*' or '*sām*' is followed by a word
//starting with '*rāj*' or '*rāṭ*' or '*ṛāñ*', then the *m* remains
//unchanged.

```
if X ∈ {sam, sām} ∧ Y ∈ {rāj, rāṭ, ṛāñ}
{
    //Skip Rule 35
    Continue processing from Rule 36;
}
```

//35. *mo'nusvāraḥ* || 8.3.23 ||
//Common Name: *anusvāra sandhi*
//If *m* at the end of a word is followed by any consonant,
//then *m* is replaced by *anusvāra*.

```
if x₁(30) ∧ y₁(2)
{
    x₁(43) = 1;
}
```

//36. *ṅaṇoḥ kuk ṭuk śari* || 8.3.28 ||
//If *ṅ* or *ṇ* is followed by a sibilant, then *k* or *ṭ* is optionally
//added respectively.

```
if x₁(28) ∧ (y₁(23) ∨ y₁(24) ∨ y₁(25))
{
    Add X|Y to word_list;
    z₁(36) = 1;
    z₂ = x₂;
    Add z to the end of X;
}
```

//37. *ḍaḥ si dhuṭ* || 8.3.29 ||
//Common name: *dhuḍāgama sandhi*
//If *ḍ* is followed by *s*, then *dh* is added optionally

```
if (x₁(38) ∧ x₂(1)) ∧ y₁(24)
{
    Add X|Y to word_list;
    z₁(34) = 1;
    z₂ = x₂;
    Add z to the end of X;
}
```

//38. *naśca* || 8.3.30 ||
//Common name: *dhuḍāgama sandhi*
//If *n* is followed by *s*, then *dh* is added optionally.

```
if (x₁(27) ∧ x₂(2)) ∧ y₁(24)
{
    Add X|Y to word_list;
    z₁(39) = 1;
    z₂ = x₂;
    Add z to the end of X;
}
```

//39. *śi tuk* || 8.3.31 ||
//Common name: *tugāgama sandhi*
//If *n* is followed by *ś*, then *t* is optionally added.

```
if (x₁(27) ∧ x₂(2)) ∧ y₁(25)
{
    Add X|Y to word_list;
    z₁(36) = 1;
    z₂ = x₂;
    Add z to the end of X;
}
```

//40. *ṅamo hrasvādaci ṅamuṇnityaṁ* ||8.3.32||
//Common name: *ṅamuḍāgama sandhi*
//If *ṅ*, *ṇ* or *n* is preceded by a short vowel and succeeded by
//a vowel, then the *ṅ*, *ṇ* or *n* gets duplicated.

*if* $(x_1(27) \land (x_2(0) \lor x_2(1) \lor x_2(2))) \land u_1(8) \land y_1(0)$
{
   Add another *x* to the end of *X*;
}

//41. *śarpare visarjanīyaḥ* || 8.3.35 ||
//If *visarga* is followed by a hard consonant which is in turn
//followed by a sibilant, then the *visarga* is retained.
*if* $x_1(44) \land (y_1(36) \lor y_1(37) \lor y_1(23) \lor y_1(24) \lor y_1(25)) \land (w_1(23) \lor w_1(24) \lor w_1(25))$
{
   Store result with no change
}

//42. *vā śari* || 8.3.36 ||
//If *visarga* is followed by a sibilant, then the *visarga* is
//optionally retained.
*if* $x_1(44) \land (y_1(23) \lor y_1(24) \lor y_1(25))$
{
   Add *X|Y* to *word_list*;
   delete *x*;
}

//43. *kupvoḥ ⋋k ⋎pau ca* || 8.3.37 ||
//If *visarga* is followed by a hard guttural or hard labial,
//then it is replaced optionally by ⋋ (pronounced as at the
//end of '*kah*') or ⋎ (pronounced as at the end of '*paf*')
*if* $x_1(44) \land y_1(41)$
{
   Add *X|Y* to word_list;
   $x_1(42) = 1$;
   $x_2 = y_2$;
   Flag_kupvoh_sutra_fired = true //flag is set
}

//44. *so'padādau* || 8.3.38 ||
//If *visarga* is followed by *pāśa, kalpa, ka* or *kāmya*, then
//the *visarga* is replaced by *s*.
*if* $x_1(44) \land Y$ begins with {*pāśa, kalpa, ka, kāmya*}
{
   $x_1(24) = 1$;
}

//45. *iṇaḥ ṣaḥ* || 8.3.39 ||
//If *visarga* is preceded by '*i*' or '*u*' and followed by *pāśa*,
//*kalpa, ka* or *kāmya*, then the *visarga* is replaced by *ṣ*.
*if* $x_1(44) \land u_1(6) \land Y$ begins with {*pāśa, kalpa, ka, kāmya*}
{
   $x_1(23) = 1$;
}

//46. *namaspurasorgatyoḥ* || 8.3.40 ||
//If '*namaḥ*' or '*puraḥ*' is followed by a hard guttural or
//hard labial, then the *visarga* is replaced by *s* optionally.
*if* $X \in$ {*namaḥ, puraḥ*} $\land y_1(41)$
{
   Add *X|Y* to *word_list*;
   $x_1(24) = 1$;
}

//47. *idudupadhasya cā'pratyayasya* || 8.3.41||
//If *visarga* is preceded by '*i*' or '*u*' and is at the end of any
//of *niḥ, duḥ, bahiḥ, āviḥ, catuḥ, prāduḥ* and is followed by
//a hard guttural or hard labial, then the *visarga* is replaced
//by *ṣ*.
*if* $X \in$ {*niḥ, duḥ, bahiḥ, āviḥ, catuḥ, prāduḥ*} $\land y_1(41)$
{
   $x_1(23) = 1$;
}

//48. *tiraso'nyatarasyām* || 8.3.42 ||
//If the word '*tiraḥ*' is followed by a hard guttural or hard
//labial, then the *visarga* is optionally replaced by *s*.
*if* $X =$ '*tiraḥ*' $\land y_1(41)$
{
   Add *X|Y* to the word_list;
   $x_1(24) = 1$;
}

//49. *dvistriścaturiti kṛtvorthe* || 8.3.43 ||
//If the words *dviḥ, triḥ* or *catuḥ* are followed by a hard
//guttural or hard labial, then the *visarga* is optionally
//replaced by *ṣ*.
*if* $X \in$ {*dviḥ, triḥ, catuḥ*} $\land y_1(41)$
{
   Add *X|Y* to the word_list;
   $x_1(23) = 1$;
}

//50. *ataḥ kṛkamikaṁsakumbhapātrakuśā*
//*karṇīṣvanavyayasya* || 8.3.46 ||
//If *visarga* is preceded by *a* and followed by a form of *kṛ*
//or *kam* or by the words *kaṁsa, kumbha, pātra, kuśā* or
//*karṇī*, then the *visarga* is replaced by *s*.
*if* $x_1(44) \land u_1(4) \land Y$ begins with {*kṛ, kar, kur, kam, kām, kaṁsa, kumbha, pātra, kuśā, karṇī*}
{
   $x_1(24) = 1$;
}

//51. *adhaḥ śirasī pade* || 8.3.47 ||
//If the word *adhaḥ* or *śiraḥ* is followed by the word '*pada*',
//then the *visarga* is replaced by *s* optionally
*if* $X \in$ {*adhaḥ, śiraḥ*} $\land Y$ begins with {*pad*}
{
   Add *X|Y* to *word_list*;
   $x_1(24) = 1$;
}

//Rules 41 to 51 form exceptions to the following rule, Rule
//52, and hence have been handled before it.

//52. *visarjanīyasya saḥ* || 8.3.34 ||
//If *visarga* is followed by a hard consonant, then *s* replaces
//the *visarga*.
$if$ Flag_kupvoh_sutra_fired = false //8.3.37 is not fired
{
   $if$ $x_1(44) \wedge (y_1(36) \vee y_1(37) \vee y_1(23) \vee y_1(24) \vee y_1(25))$
   {
     $x_1(24) = 1;$
   }
}

//53. *raṣābhyāṁ no ṇaḥ samānapade* || 8.4.1 ||
//   *aṭkupvāṅnumvyavāye'pi* || 8.4.2 ||
//   *padāntasya* || 8.4.37 ||
//If *n* is preceded by *ṛ, ṝ, r* or *ṣ* within the same word, and a
//palatal, cerebral, dental, *l, ś* or *s* does not lie between the
//two, and *n* is not the last letter of the word, then *n* is
//replaced by *ṇ*.
$for$ (each letter $y$ in $X$ where $y$ is not the last letter)
{
   $if$ $(y_1(27) \wedge y_2(2))$ //*y* is 'n'
   {
     $if$ $\exists$ a letter $x$ in $X$ preceding $y$ where $((x_1(12) \wedge \neg x_2(2)) \vee x_1(21) \vee x_1(23))$ //*x* is *ṛ, ṝ, r* or *ṣ*
     {
       $if$ $\nexists$ any letter $q$ between $x$ and $y$ where $q_1(32) \vee q_1(33) \vee q_1(34) \vee (q_1(20) \wedge q_2(3)) \vee q_1(24) \vee q_1(25)$
       {
         $y_2(1) = 1;$
       }
     }
   }
}
//Repeat the above for Y

//54. *stoḥ ścunāḥ ścuḥ* || 8.4.40 ||
//   *śāt* || 8.4.44 ||
//Common name: *ścutva sandhi*
//If a palatal other than *ś* is followed by a dental, or a dental
//is followed by a palatal, then the dental is replaced by the
//corresponding palatal.
$if$ $(x_1(32) \wedge \neg x_2(5)) \wedge (y_1(34))$
{
   $y_1(32) = 1;$
}
$elseif$ $x_1(34) \wedge y_1(32)$
{
   $x_1(32) = 1;$
}

//55. *ṣṭunāḥ ṣṭuḥ* || 8.4.41 ||
//   *na padāntāṭṭoranām* || 8.4.42 ||
//   *toḥ ṣi* || 8.4.43 ||
//Common name: *ṣṭutva sandhi*
//If a dental is followed by a cerebral except *ṣ*, or if the
//specific cerebral *ṭ* is followed by *nām, navati* or *nagarī*, or
//if a cerebral is followed by a dental, then the dental is
//replaced by the corresponding cerebral.
$if$ $(x_1(34) \wedge \neg x_2(5)) \wedge (y_1(33))$
{
   $x_1(33) = 1;$
}
$elseif$ $(x_1(33) \wedge x_2(0)) \wedge Y \in \{nām, navat, nagar\}$
{
   $y_1(33) = 1;$
}
$elseif$ $x_1(33) \wedge y_1(34)$
{
   $y_1(33) = 1;$
}

//56. *yaro'nunāsike'nunāsiko vā* || 8.4.45 ||
//Common name: *anunāsikā sandhi*
//If a consonant other than '*h*' is followed by a nasal, then
//the consonant is optionally replaced by the corresponding
//nasal. The rule is obligatory if the second word is
//'*mayam*' or '*mātram*'.
$if$ $(x_1(36) \vee x_1(37) \vee x_1(38) \vee x_1(39))$
{
   $if$ $y_1(27)$
   {
     Add $X|Y$ to the word_list;
     $x_1(27) = 1;$
   }
   $elseif$ $Y \in \{maya, mātra\}$
   {
     $x_1(27) = 1;$
   }
}

//57. *aco rahābhyāṁ dve* || 8.4.46 ||
//If *r* or *h* is followed by any consonant other than *h* and
//preceded by a vowel, then the consonant is duplicated
//within a word.
$for$ (each set of consecutive letters $u,x,y$ in $X$)
{
   $if$ $(x_1(21) \vee x_1(26)) \wedge (y_1(2) \wedge \neg y_1(26)) \wedge u_1(0)$
   {
     $z = y;$
     Add $z$ after $x;$
   }
}
$for$ (each set of consecutive letters $u,x,y$ in $Y$)
{
   $if$ $(x_1(21) \vee x_1(26)) \wedge (y_1(2) \wedge \neg y_1(26)) \wedge u_1(0)$
   {
     $z = y;$
     Add $z$ after $x;$
   }
}

//58. *anaci ca* || 8.4.47 ||
//     *dīrghādācāryāṇām* || 8.4.52 ||
//If any consonant other than *h* is preceded by a short vowel
//and followed by anything other than a vowel, then the
//consonant is doubled within a word.
$for$ (each set of consecutive letters $x,y,w$ in $X$)
{
   $if\ x_1(8) \land (y_1(2) \land \neg y_1(26)) \land (w_1(1) \lor w_1(2) \lor$
   $w_1(3))$
   {
     $z = y$;
     Add $z$ after $x$;
   }
}
$for$ (each set of consecutive letters $x,y,w$ in $Y$)
{
   $if\ x_1(8) \land (y_1(2) \land \neg y_1(26)) \land (w_1(1) \lor w_1(2) \lor$
   $w_1(3))$
   {
     $z = y$;
     Add $z$ after $x$;
   }
}

//59. *jhalām jaś jhaśi* || 8.4.53 ||
//Common name: *jaśtva sandhi*
//If a non-nasal mute, sibilant or aspirate is followed by
//Column3 or Column4, then the first letter is replaced by
//the corresponding Column3 letter.
$for$ (each set of consecutive letters $x, y$ in $X$)
{
   $if\ (y_1(38) \lor y_1(39))$
   {
     $if\ x_1(36) \lor x_1(37) \lor x_1(38) \lor x_1(39) \lor x_1(26)$
     {
       $x_1(38) = 1$;
     }
     $elseif\ x_1(23) \lor x_1(24) \lor x_1(25)$
     {
       $x_1(38) = 1$;
       $if\ x_1(23)$
       {
          $x_2(1) = 1$;
       }
       $elseif\ x_1(24)$
       {
          $x_2(2) = 1$;
       }
       $elseif\ x_1(25)$
       {
          $x_2(3) = 1$;
       }
     }
   }
}
//Repeat the above for Y

//60. *khari ca* || 8.4.55 ||
//Common name: *cartva sandhi*
//If a non-nasal mute or sibilant is followed by a
//hard consonant, then the first letter is replaced by the
//corresponding Column1 letter.
$if\ (x_1(36) \lor x_1(37) \lor x_1(38) \lor x_1(39) \lor x_1(23) \lor$
$x_1(24) \lor x_1(25)) \land (y_1(36) \lor y_1(37) \lor y_1(23) \lor$
$y_1(24) \lor y_1(25))$
{
   $x_1(36) = 1$;
}

//Check internally in each word too
$for$ (each set of consecutive letters $x, y$ in $X$)
{
   $if\ y_1(36) \lor y_1(37) \lor y_1(23) \lor y_1(24) \lor y_1(25)$
   {
     $if\ (x_1(36) \lor x_1(37) \lor x_1(38) \lor x_1(39))$
     {
       $x_1(36) = 1$;
     }
     $elseif\ x_1(23) \lor x_1(24) \lor x_1(25)$
     {
       $x_1(36) = 1$;
       $if\ x_1(23)$
       {
          $x_2(1) = 1$;
       }
       $elseif\ x_1(24)$
       {
          $x_2(2) = 1$;
       }
       $elseif\ x_1(25)$
       {
          $x_2(3) = 1$;
       }
     }
   }
}
//Repeat the above for *Y*

//61. *anusvārasya yayi parasavarṇaḥ* ||8.4.58||
//Common name: *parasavarṇa sandhi*
//If *anusvāra* is followed by a semi-vowel or a mute, then
//the *anusvāra* is replaced by the nasal equivalent of the
//second letter.
$if\ x_1(43)$
{
   $if\ y_1(20)$
   {
     $x_1(22) = 1$;
     $x_2 = y_2$;
   }
   $elseif\ y_1(36) \lor y_1(37) \lor y_1(38) \lor y_1(39)$
   {
     $x_1(27) = 1$;
     $x_2 = y_2$;

```
    }
}

//62. torli || 8.4.60 ||
//Common name: parasavarṇa sandhi
//i) If n is followed by l, then n is replaced by nasal l.
//ii) If a dental other than n and s is followed by l, then
//the dental is replaced by l.
if (x_1(27) ∧ x_2(2)) ∧ (y_1(20) ∧ y_2(3))
{
    x_1(22) = 1;
    x_2 = y_2;
}
elseif (x_1(34) ∧ ¬(x_2(4) ∨ x_2(5))) ∧ (y_1(20) ∧ y_2(3))
{
    x = y;
}

//63. jhayoho'nyatarasyām || 8.4.62 ||
//Common name: pūrvasavarṇa sandhi
//If a non-nasal mute is followed by h, then h is optionally
//replaced by the Column4 letter corresponding to the non-
//nasal mute.
if (x_1(36) ∨ x_1(37) ∨ x_1(38) ∨ x_1(39)) ∧ y_1(26)
{
    Add X|Y to the word_list;
    y_1(39) = 1;
    y_2 = x_2;
}

//64. śaścho'ṭi || 8.4.63 ||
//Common name: chatva sandhi
//If a non-nasal mute is followed by ś which is in turn
//followed by a vowel, semi-vowel or nasal, then ś is
//optionally replaced by ch.
if(x_1(36) ∨ x_1(37) ∨ x_1(38) ∨ x_1(39)) ∧ y_1(25) ∧
(w_1(0) ∨ w_1(1) ∨ w_1(27))
{
    Add X|Y to the word_list;
    y_1(40) = 1;
}
//65. halo yamāṁ yami lopaḥ || 8.4.64 ||
//If a semi-vowel or nasal is preceded by a consonant and
//followed by the same semi-vowel or nasal letter, then one
//of the duplicate letters is dropped.
if u_1(2)
{
    if (x_1(20) ∨ x_1(27)) ∧ y == x
    {
        delete x;
    }
}

//66. jharo jhari savarṇe || 8.4.65 ||
//If a non-nasal mute or sibilant is preceded by a consonant
//or semi-vowel and followed by a homogeneous mute or
//sibilant, then one of the duplicate letters is optionally
//dropped, within a word
for (each set of consecutive letters u,x,y in X)
{
    if u_1(1) ∨ u_1(2)
    {
        if (x_1(36) ∨ x_1(37) ∨ x_1(38) ∨ x_1(39) ∨
        x_1(23) ∨ x_1(24) ∨ x_1(25)) ∧ (x_1 == y_1)
        {
            Add X|Y to the word_list;
            delete x;
        }
    }
}
```
// Repeat the above for Y

*C. The Search Engine*

Algorithm SandhiProcessor may be used to generate alternative forms of a given search word. The following algorithm is used to generate all possible alternative forms of a given word, by providing possible word forms before and after the word so that *sandhi* rules get triggered. All these word forms are searched for in the E-text.

Algorithm GenerateAllWordForms ($Z$)
```
{
//Z is the search word.
//{WordForms} denotes the set of word forms generated by
//the algorithm and is initially the null set.
    Add Z to {WordForms};
    X = Z;
    for (each y in {vowels, semi-vowels, consonants})
    {
        Add SandhiProcessor(X, Y) to {WordForms};
    }

    for (each Y ∈ {WordForms})
    {
        for (each X in {vowels, semi-vowels, consonants, ḥ, ṁ,
        #})
        {
            Add SandhiProcessor(X, Y) to {WordForms};
        }
    }
}
```

## VI. CONCLUSION

The schema developed in this work presents a simple yet unique and efficient method to process the *sandhi* aphorisms of Pāṇini. The letter representation scheme is binary, and hence all the checks are implemented as bit-level operations and simple bit-set and bit-unset operations suffice to carry out the *sandhi* transformation. The efficiency is further enhanced by the division of a letter representation into two parts and the consequent reduction of the transformation process to a shifting of category. Further, this pattern of solving the sandhi construction problem is unprecedented in the literature. Thus, representation schema and the results of the *sandhi*-processing algorithm represent an efficient computational model to process Sanskrit euphonic conjunctions. It must be mentioned here that some rules such as those with regard to *prakṛtibhāva sandhi* (non-transformational *sandhi*) have not been presented above. However, it is clear that their implementation is only an extension of the algorithm presented in this work that does not require any new schema.

The representational schema has been reduced further from [4] in this work, since many more *sandhi* rules have been incorporated here. The optimality of the schema is clear from the simplicity of the rule representation.

The final algorithm presented in this work, which uses this *sandhi* processor for word searches in E-texts is the first of its kind in the literature with regard to Sanskrit.

The use of the *sandhi* processor for searching ensures comprehensiveness of the search, while the efficiency of the *sandhi* processing method presented in this work ensures that search speeds are not compromised due to the increase in the number of words to be searched for. This was confirmed in the implementation of the algorithm.

The algorithms presented in this work have been tested for use with Sanskrit E-text in *Devanāgarī* script after a conversion engine converted *Devanāgarī* Unicode to Latin Unicode E-text.